\documentclass[letterpaper]{article} 
\usepackage{aaai23}  
\nocopyright
\usepackage{times}  
\usepackage{helvet}  
\usepackage{courier}  
\usepackage[hyphens]{url}  
\usepackage{graphicx} 
\usepackage{natbib}  
\usepackage{caption} 
\usepackage{algorithm}
\usepackage{algorithmic}
\usepackage{booktabs}
\usepackage{multirow}
\usepackage{amsmath}
\usepackage{adjustbox}

\usepackage{newfloat}
\usepackage{listings}
\DeclareCaptionStyle{ruled}{labelfont=normalfont,labelsep=colon,strut=off} 
\floatstyle{ruled}
\newfloat{listing}{tb}{lst}{}
\floatname{listing}{Listing}
\title{Auditing Algorithmic Fairness in Machine Learning for Health with Severity-based LOGAN}
\author{
    Anaelia Ovalle,
    Sunipa Dev, 
    Jieyu Zhao,
    Majid Sarrafzadeh,
    Kai-Wei Chang
}
\affiliations{
    University of California, Los Angeles\\

    \{anaelia, sunipa\}@cs.ucla.edu, jieyuzhao@ucla.edu,\{majid, kwchang\}@cs.ucla.edu
}

\usepackage{bibentry}
\begin{document}

\maketitle

\begin{abstract}
Auditing machine learning-based (ML) healthcare tools for bias is critical to preventing patient harm, especially in communities that disproportionately face health inequities. General frameworks are becoming increasingly available to measure ML fairness gaps between groups. However, ML for health (ML4H) auditing principles call for a contextual, patient-centered approach to model assessment. Therefore, ML auditing tools must be (1) better aligned with ML4H auditing principles and (2) able to illuminate and characterize communities vulnerable to the most harm. To address this gap, we propose supplementing ML4H auditing frameworks with SLOGAN (patient Severity-based LOcal Group biAs detectioN), an automatic tool for capturing local biases in a clinical prediction task. SLOGAN adapts an existing tool, LOGAN (LOcal Group biAs detectioN), by contextualizing group bias detection in patient illness severity and past medical history. We investigate and compare SLOGAN's bias detection capabilities to LOGAN and other clustering techniques across patient subgroups in the MIMIC-III dataset. On average, SLOGAN identifies larger fairness disparities in over 75\%  of patient groups than LOGAN while maintaining clustering quality. Furthermore, in a diabetes case study, health disparity literature corroborates the characterizations of the most biased clusters identified by SLOGAN. Our results contribute to the broader discussion of how machine learning biases may perpetuate existing healthcare disparities.


\end{abstract}

\section{Introduction}

Fairness auditing frameworks are necessary for operationalizing machine learning algorithms in healthcare (ML4H). In particular, they must identify and characterize biases \citep{chen2021ethical}. Ongoing directives to promote health equity must also translate to these spaces, with care placed on those historically vulnerable to the most harm, such as communities with chronic illnesses and racial and ethnic minorities \citep{oala2020ml4h,laura_joszt_2022}. To do this, they must be prioritized when evaluating for fairness in ML4H \citep{ rajkomar2018ensuring, chen2021ethical,roosli2022peeking}.\looseness=-1

Commercialized auditing tools are being increasingly leveraged for bias assessment in ML4H algorithms \citep{oala2020ml4h, kumar2020machine}. However, we argue that applying out-of-the-box auditing tools without a clear patient-centric design is not enough. Existing auditing tools must align with health ethics principles that guide a framework's operationalization. In guiding ML4H auditing literature, this means the tool must be able to detect locally biased patient subgroups when monitoring the fairness of ML4H throughout its lifecycle \citep{de2022guidelines}. To monitor disparities with health equity in mind, researchers must also engage critically with the broader sociotechnical context surrounding the use of ML auditing tools in healthcare \citep{pfohl2021empirical}. \looseness=-1


This work addresses the gap by devising a patient-centric ML auditing tool called SLOGAN. SLOGAN adapts LOGAN \citep{zhao2020logan}, an unsupervised algorithm that uses contextual word embeddings \citep{devlin2018bert} to cluster local groups of bias indicated by model performance differences. To better align auditing with measures of effective care planning and therapeutic intervention \citep{katz2016genesis}, SLOGAN identifies local group biases in clinical prediction tasks by leveraging patient risk stratification. Previous medical history is also commonly used for understanding health inequities through social, cultural, and structural barriers the patient experiences \citep{brennan2008promoting}. Therefore, SLOGAN characterizes these local biases using patients' electronic healthcare records (EHR) histories. \looseness=-1

Experiments on in-hospital mortality prediction demonstrate how SLOGAN effectively identifies local group biases. We audit the model across 12 MIMIC-III patient subgroups. We then provide a case study to further examine fairness differences in patients with chronic illnesses such as Diabetes Mellitus. Results indicate that (1) SLOGAN, on average, captures more considerable biases than LOGAN, and (2) such identified biases align with existing health disparity literature.

\section{Background and Related Work}\label{background}

\subsection{Algorithmic Auditing in ML for Healthcare}
\citet{obermeyer2019dissecting} audit a commercialized ML4H algorithm by dissecting observed disparities between patient risk and overall health cost. The authors call for the continued probing of health inequity in these clinical systems. Likewise, \citet{wiegand2019and, pfohl2021empirical,siala2022shifting, de2022guidelines} create guidelines for operationalizing transparent assessments of ML4H models. Auditing frameworks such as Aequitas \footnote{http://aequitas.dssg.io/} and AIFairness360 \footnote{https://aif360.mybluemix.net/} are operationalized for this purpose \citep{oala2020ml4h}. The tools provide reports relevant to protected groups and fairness metrics, indicating unfairness through preset disparity ranges. 
 
\subsection{Measuring Health Equity Barriers}
Intersectional social identities are related to a patient's health outcomes \citep{mcginnis2002case, katz2018association}. Therefore, measuring health equity in ML requires understanding a patient beyond their illness. In practice, this can include focusing on populations with histories of a significant illness burden or examining bias from the lens of social determinants of health (SDOH). Fairness literature has also dictated a need to measure biases from multidimensional perspectives \citep{hanna2020towards}. Capturing social context beyond protected attributes is helpful for this cause. SDOH, such as unequal access to healthcare, language, stigma, racism, and social community, are underlying contributing factors to health inequities \citep{aday1994health, peek2007diabetes,brennan2008promoting}.

\subsection{Fairness and Local Bias Detection}
LOGAN \citep{zhao2020logan}, a method to detect local bias, adapts K-Means to cluster BERT embeddings while maximizing a bias metric within each cluster. 
LOGAN
consists of a 2-part objective:  a K-Means clustering objective ($L_c$)  and an objective to maximize a bias metric ($L_b$, e.g. the performance gap between 2 groups) within each respective cluster.\looseness=-1
\setlength{\abovedisplayskip}{0.0pt}
\setlength{\belowdisplayskip}{0.0pt}
\begin{equation}
\min_C \quad L_c+\lambda L_b
\label{eq:1}
\end{equation}
where 
$\lambda \leq 0$ is a tunable hyperparameter to control the tradeoff between the two objectives and indicates how strongly to cluster with respect to group performance differences. We define our bias metric as the model performance disparity between 2 groups, measured by accuracy. However, detecting biases by identifying similar contextual representations is not enough. The task must be adapted to the clinical domain to audit with health equity in mind. One way to do this is by incorporating domain-specific information. For example, severity scores stratify patients based on their immediate needs and help clinicians decide how to allocate resources effectively. Therefore, we build off of LOGAN and create a tool that translates to the medical setting by mindfully using this information \cite{ferreira2001serial}.


\section{Methodology}


\subsection{Clinical NLP Pretrained Embeddings}
Several BERT models are publicly available for use in the clinical setting. These include various implementations of ClinicalBERT \citep{alsentzer-etal-2019-publicly, huang2019clinicalbert}. We proceed with leveraging a variant of ClinicalBERT from \citet{zhang2020hurtful} as this is an extension of ClinicalBERT with improvements such as whole-word masking.\looseness=-1

\begin{table*}[t]
\centering
\begin{tabular}{llll|l} 
\toprule
                         & K-Means  & LOGAN & SLOGAN & \# of MIMIC-III Attributes~  \\ 
\midrule
Inertia  ($\downarrow$)                & 1.0                & 0.991            & \textbf{0.981}    & 7/12 (58\%)                  \\ 
SCR ($\uparrow$)                      & 15.3               & 22.9             & \textbf{30.1}     & 12/12 (100\%)                \\ 
SIR   ($\uparrow$)                    & 15.3               & 18.4             & \textbf{23.4}     & 7/12 (58\%)                  \\ 
\textbar{}Bias\textbar{} ($\uparrow$) & 12.5               & 21.5             & \textbf{34.2}     & 9/12 (75\%)                  \\
\bottomrule
\end{tabular}
\caption{Average values for 12 MIMIC-III attributes across models and evaluation metrics. SCR, SIR, and \textbar{}Bias\textbar{} in \%. \textbar{}Bias\textbar{} is the average absolute model performance difference in biased clusters. Bold is the best performance per row. Right-most column is number of MIMIC-III attributes where SLOGAN performs best. Arrows indicate desired direction of a number.}
\label{tbl:avg_bias}
\end{table*}


\subsection{Automatic Bias Detection} To create a patient-centric bias detection tool, we encourage SLOGAN to identify large bias gaps while accounting for similarity in patient severity. SLOGAN measures local biases in a model using patient-specific features and contextual embeddings of patient history for in-hospital mortality prediction. We do this via a patient similarity constraint. A variety of patient severity scores such as OASIS, SAPS II, and SOFA are available for use \citep{le1993new, jones2009sequential, johnson2013new}. Following health literature and clinician advice, we select the SOFA acuity score. However, depending on clinician needs, a different constraint may be used (e.g., ICD-9 codes). Extending Eq. ~\eqref{eq:1}, this results in the following optimization problem:

\setlength{\abovedisplayskip}{0.0pt}
\setlength{\belowdisplayskip}{0.0pt}
\begin{equation}
\min_C \quad L_c+\lambda L_b  + \gamma L_s
\end{equation}
where $L_s$ is added to encourage the model to group patients with  similar acute severity. $\lambda \leq 0$ and $\gamma \geq 0$ are hyperparameters that control the tradeoff between the objectives of grouping patient similarity and clustering by local bias.

\setlength{\abovedisplayskip}{0.0pt}
\setlength{\belowdisplayskip}{0.0pt}
\begin{equation}
L_s = \sum_{j=1}^{k} \Bigg|\sum_{x_i \in A} SOFA_{ij} - \sum_{x_i \in B} SOFA_{ij}\Bigg|^2
\end{equation}
$\lambda$ and $\gamma$ are tuned via a grid search and we choose the combination that identifies the largest local group biases (Appendix Table \ref{tbl:appendix_params}).\looseness=-1

We define the bias score as having at least a 10\% difference in accuracy and at most a SOFA score difference of 0.8. ~\footnote{We choose the thresholds by splitting the data and creating bootstrap estimates 1000 times, then add three standard deviations.} We compare SLOGAN to LOGAN and K-Means across three metrics. To measure the utility of the clusters found, we examine the ratio of biased clusters found (SCR) and the number of instances in those clusters (SIR). We use inertia to measure clustering quality, as it reflects how well the data clustered across respective centroids. Finally, we compare each algorithm's inertia to a baseline K-Means model normalized to 1.0.\looseness=-1

\subsection{Data and Setup} In order to maximize reproducibility, we perform experiments with the same patient cohorts defined in the benchmark dataset from the MIMIC-III clinical database \citep{johnson2016mimic,harutyunyan2019multitask}. Following  \citet{sun2022negative}, to understand how BERT represents social determinants of health and captures possible stigmatizing language in the data, we extracted the history of present illness, past medical history, social history, and family history across physicians, nursing, and discharge summaries \citep{marmot2005social}. We employed MedSpacy \citep{eyre2021medspacy} to extract any information related to a patient's social determinants of health. After preprocessing, this translated into a 70\% train, 15\% validation, and 15\% test split of 1581, 393, and 309 patients, respectively. No patient appeared across the splits. Analyses were conducted across self-identified ethnicity, sex, insurance type, English speaking, presence of chronic illness, presence of diabetes (type I and II), social determinants of health, and negative patient descriptors to measure stigma. We also explored creating cross-sectional groups (Appendix Table \ref{tbl:appendix_attributes}).\looseness=-1

We used SLOGAN to audit a fully connected neural network from \citet{zhang2020hurtful} used to predict in-hospital mortality, a common MIMIC-3 benchmarking task \cite{harutyunyan2019multitask}. \footnote{A patient that has passed within 48 hours of their ICU stay is assigned the label of 1, otherwise patients are assigned the label 0.} Each patient note in the test set was encoded and concatenated with gender, OASIS, SAPS II, SOFA scores, and age. To provide a rich contextual representation of patient notes to SLOGAN, encodings consisted of the concatenated last four layers of ClinicalBERT \citep{devlin2018bert}. The embeddings encoded 512 tokens, the maximum number of tokens for BERT. We followed the best hyperparameters of the model and chose the threshold that provides at least 80\% accuracy on the validation set.\looseness=-1 
\vspace{-0.5mm}


\section{Results}
\vspace{-2mm}

\subsection{Aggregate Analysis} 
We assessed SLOGAN's local bias clustering abilities and quality across 12 attributes in MIMIC-III, including demographic variables such as ethnicity and gender. The model was compared to K-Means and LOGAN using the SCR, SIR, $|$Bias$|$, and Inertia measurements introduced in the previous sections. We report these results in Table \ref{tbl:avg_bias}. In most attributes, SLOGAN was the best at identifying groups with fairness gaps. Identified groups contained more instances and larger biases, while maintaining clustering quality. In particular, SLOGAN identified the most and largest local group biases in at least 9/12 (75\%) attributes, measured by SCR and $|$Bias$|$, respectively. When comparing LOGAN and K-Means, SLOGAN found the highest ratio of biased instances within biased clusters (SIR) in 7/12 (58\%) MIMIC-3 attributes. We report audits across all attributes in Appendix Table \ref{tbl:appendix_comparison}. \looseness=-1 




\subsection{Case Study: Diabetes Mellitus} 

\begin{table}[h]
\small
\centering
\begin{tabular}{llllll} 
\toprule
\multirow{5}{*}[-0.4em]{Has Diabetes} & Method  & Acc-Yes & Acc-No &  \textbar{}Bias\textbar{}          \\ 
\cmidrule{2-5}
                              & Global  & 75.0    & 84.1   & 9.1            \\
                              & K-Means & 55.0    & 75.0   & 20.0           \\
                              & LOGAN   & 60.0    & 88.0   & 28.0           \\
                              & SLOGAN  & 54.5    & 91.7   & \textbf{37.1}  \\
\bottomrule
\end{tabular}

\caption{Bias detection (\%) for in-hospital mortality task. Global indicates global bias. ``Yes'' indicates patient with diabetes. \textbar{}Bias\textbar{} is the max absolute model performance difference in biased clusters. SLOGAN identifies local biases greater than global bias observed in the data (bold). }
\label{tbl:max_bias}
\end{table}

\begin{table}[h]
\centering
\small
\begin{tabular}{llllll} 
\toprule
\multirow{4}{*}[-0.4em]{Has Diabetes} & Method  & Inertia & SCR  & SIR  & \textbar{}Bias\textbar{}  \\ 
\cmidrule{2-6}
                              & K-Means & 1.00    & 33.3 & 27.1 & 14.2                      \\
                              & LOGAN   & 1.003   & 25.0 & 16.9 & 25.0                      \\
                              & SLOGAN  & 1.12    & 25.0 & 15.4 & \textbf{28.6}             \\
\bottomrule
\end{tabular}
\caption{Comparison under diabetes attribute. SCR and SIR are respectively the \% of biased clusters and \% of biased instances. \textbar{}Bias\textbar{}(\%) is the average absolute bias score for the biased clusters. SLOGAN finds the largest bias (bold).}
\label{tbl:scr_sir}
\vspace{-10pt}
\end{table}



\subsubsection{Cluster Analysis}
Diabetes is one of the most common and costly chronic conditions worldwide, accompanied by serious comorbidities\citep{ceriello2012diabetes}. To further study this, we used SLOGAN to assess the local group biases on the \textbf{HAS DIABETES} attribute and identified fairness gaps in agreement with health literature.\looseness=-1

We report the accuracy and maximum absolute performance differences across identified biased clusters by K-Means, LOGAN, and SLOGAN in Table \ref{tbl:max_bias}. The performance difference overall between patients that do and do not have diabetes was 9.1\%. K-Means and LOGAN identified local groups with larger performance discrepancies (20\% and 28.1\%, respectively). Notably, SLOGAN performed the best at identifying a local region with the largest performance gap (37.1\%). We also report the SCR, SIR, $|$Bias$|$, and Inertia in Table \ref{tbl:scr_sir}. Results indicate that SLOGAN found groups with a larger average bias magnitude than K-Means and LOGAN. While LOGAN and SLOGAN identified the same ratio of biased clusters (25.0\%), SLOGAN identified the largest local bias region (28.6\%) with a small tradeoff in inertia (Appendix Figure \ref{app:fig:diabetes}).

\begin{figure}[t]
    \centering
    \includegraphics[width=0.8\columnwidth]{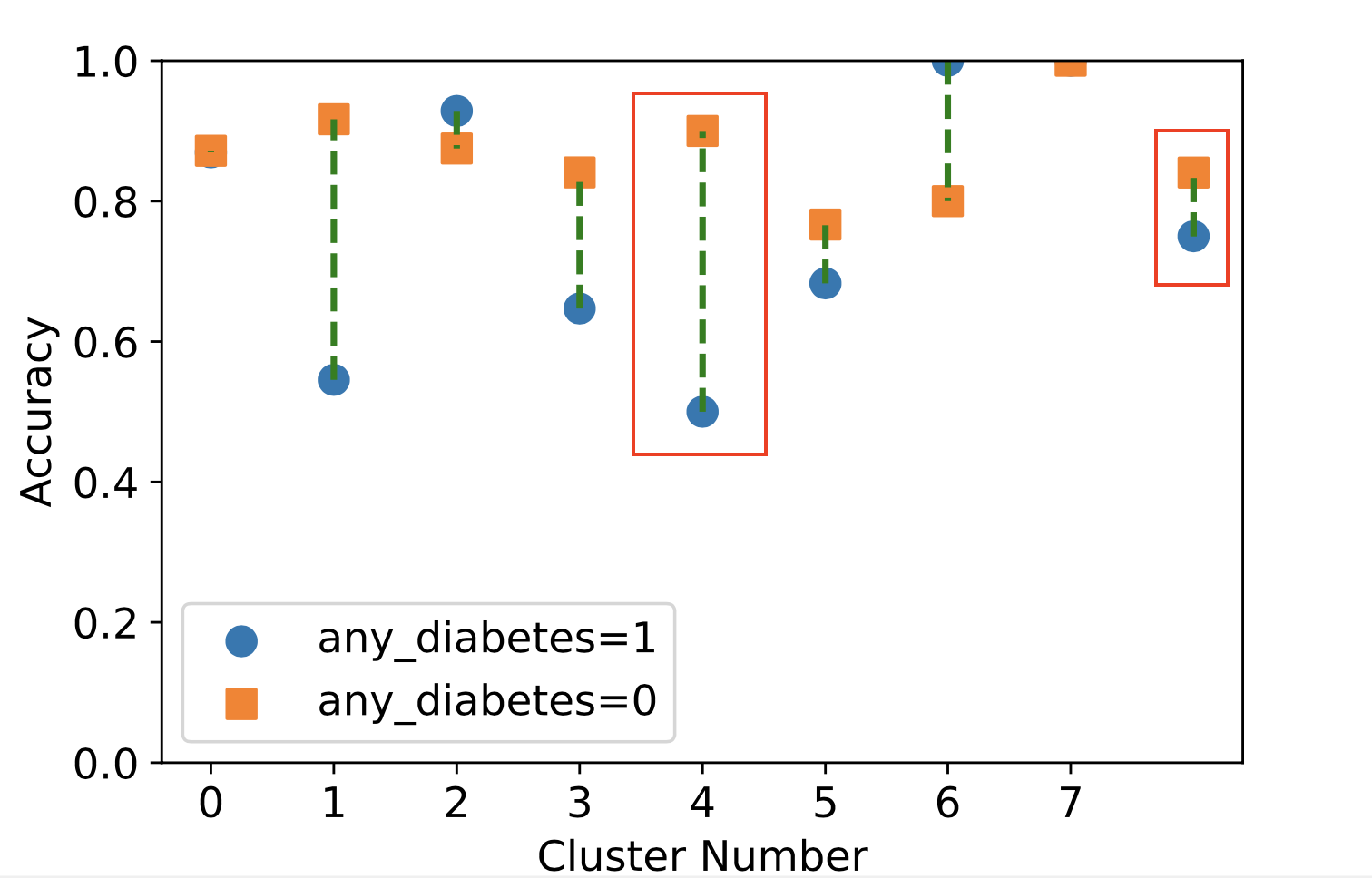}
    \caption{Performance differences for \textbf{HAS DIABETES} attribute. Furthest right red box shows global bias, while SLOGAN finds a local area of much higher bias at cluster 4. }
    \label{fig:diabetes}
\end{figure}

To more carefully examine clusters formed by SLOGAN, we show respective performance deviations in Figure \ref{fig:diabetes}. We found that SLOGAN identified fairness gaps documented in health literature. Two clusters exhibited a large local bias towards patients without diabetes, clusters 1 and 4. We analyzed differences in cluster characteristics between the most and least biased cluster. The most biased cluster, cluster 4, contained 38\% more patients with chronic illnesses besides diabetes, with 33.3\% suffering from chronic illnesses besides diabetes or hypertension. We then compared cluster 4 to all other clusters. Again, we found that it  contained the largest percentage of (1) patients (62.5\%) with chronic illnesses besides diabetes and (2) patients with chronic illnesses besides diabetes and hypertension (25\%). Cluster 4 also had fewer patients with private insurance than the least biased cluster and the lowest percentage of English-speaking patients (4.6\%) in the entire dataset (Appendix Table \ref{tbl:has_diabetes_character}). Notably, these differences in disease burden, insurance, and language align with existing research indicating how populations with the largest health disparities often suffer from a larger burden of disease and may experience significant structural language barriers \citep{flores2005impact, peek2007diabetes}. \looseness=-1


\subsubsection{Bias Interpretation with Topic Modeling}
Severe diabetes complications may result in various forms of deadly infections and respiratory issues \citep{joshi1999infections, muller2005increased,de2017type}. Provided the in-mortality task, we asked if indications of severe diabetes complications were present when using SLOGAN. To do this, we ran Latent Dirichlet Allocation topic modeling \citep{blei2003latent} within identified SLOGAN clusters. We detail the preprocessing steps in the appendix. Table \ref{tbl:lda_diabetes} lists the top 20 topic words for the most and least biased clusters. SLOGAN grouped patients with histories indicating deadly infections and respiratory issues in the most biased cluster. Terms included ``sputum'' (thick respiratory secretion), ``Acinobacter'' (bacteria that can live in respiratory secretions), and ``Vanco'' (used to treat infections).\looseness=-1

Social determinants of health also correlate to effective self-management of diabetes \citep{clark2014social,adu2019enablers}. Therefore we also examined differences in social determinants of health between the least and most biased clusters. While LDA cannot determine the directionality of SDOH impact, the top 20 terms are among the most important when forming the cluster's topic distribution. In the least biased cluster, top words included terms around the community such as `home', `offspring', `children', and `sibling'. However, in the most biased cluster, just 1 of the 20 terms, `parent', reflected possible existing social support.\looseness=-1









\section{Discussion}\label{sec:discussion}
We developed SLOGAN as a framework to audit an ML4H task by identifying areas of patient severity-aware local biases. Results indicated that SLOGAN captures more and higher quality clusters across several subgroups than the baseline models, K-Means and LOGAN. To illustrate how to use SLOGAN in a clinical context, we conducted a case study that used SLOGAN to identify clusters of local bias in diabetic patients. We found that the biases observed aligned with existing health literature. In particular, the cluster with the \textit{largest local bias} was also the cluster with the \textit{largest disease burden}. This result demonstrates a need to further examine and repeat these experiments across patient cohorts and performance metrics. Interesting future works may include asking how models encode vulnerable communities in their representations and if health disparities consistently propagate into model biases.\looseness=-1

In practice, SLOGAN can be used to determine biased clusters for review before model deployment in a healthcare setting. The tool may also track how biases shift due to changes in the data or across operationalization in different hospital networks. Furthermore, patient-centric local bias detection can supplement ML4H model auditing. With this information, ML researchers and clinicians can use auditing report cards to decide on the next steps for inclusive model development.\looseness=-1

\begin{table}[]
\centering
\small
\begin{tabular}{cp{5.5cm}}
\toprule
\multirow{5}{*}[0.2em]{\shortstack{Most biased \\ (40.0)}}                                           & parent, given, recent, vanco, treat, fever, 
acinetobacter, ecg, negative, intubated, disorder, bottles, clozaril, complete, sputum, past, started, ed, found, admitted   \\
\midrule
\multirow{5}{*}[0.2em]{\shortstack{Least biased \\ (0.2)}} & noted, past, recent, home, given, due, pain, two, offspring, mild, chest, initially, without, blood, vancomycin, children, shortness\_breath, sibling, admitted, started  \\ 
\bottomrule
\end{tabular}

\caption{Top 20 topic words in the most and least biased
clusters using SLOGAN for \textbf{HAS DIABETES} attribute. Number is the bias score (\%) of that cluster.}
\label{tbl:lda_diabetes}
\end{table}

\subsection{Ethical Statement \& Limitations}

Our analysis used MIMIC-III data, an open deidentified clinical dataset. Only credentialed researchers who fulfilled all training requirements and abided by the data use agreement accessed the data. \footnote{https://physionet.org/content/mimiciii/1.4/\#files} We review the data and clinical notes a second time to confirm the removal of any patient-related information, including location, age, name, date, or hospital. \looseness=-1

In practice, further interdisciplinary discussion on how SLOGAN can best be integrated into the ML4H auditing pipeline is welcomed. While we do not analyze the factors influencing model fairness, we encourage this future work. Furthermore, it is important to note that the absence of flagged bias clusters is not an indicator of a total absence of risk for downstream unfair outcomes.\looseness=-1

\appendix

\section{Appendix}\label{sec:appendix}

\subsection{LDA}\label{sec:appendix_lda}

LDA is run using the NLTK and gensim packages \citep{loper2002nltk, vrehuuvrek2011gensim}. Unigrams and bigrams are generated using gensim.phrases with min count=3 and threshold=5. The LDA is run on gensim with random state=100, updateevery=1, chunksize=100, and passes=100. To get achieve better topic modeling, words like child, son, daughter are tokenized as 'offspring'. Words pertaining to father or mother are replaced with 'parent'. Words such as hypertension and hypertensive are replaced with 'hypert'. Similarly, words such as hypotension and hypotensive are replaced with 'hypot'.

\subsection{Negative Patient Descriptors}
\label{sec:appendix_negative}
We explored the SDOH dimension of stigma in clinical notes through the extraction of negative patient descriptors found in \citep{sun2022negative} and outline the results in the Appendix Table \ref{tbl:lda_negative}. However, further preprocessing beyond the usage of regexes is needed to reduce false positive rates.


\subsection{Code}
\label{sec:appendix_code}
We will publically release the code in an easily accessible repository upon review of this paper.

\setcounter{table}{0}
\setcounter{figure}{0}

\begin{figure}[!h]
    \centering
    \includegraphics[width=\columnwidth]{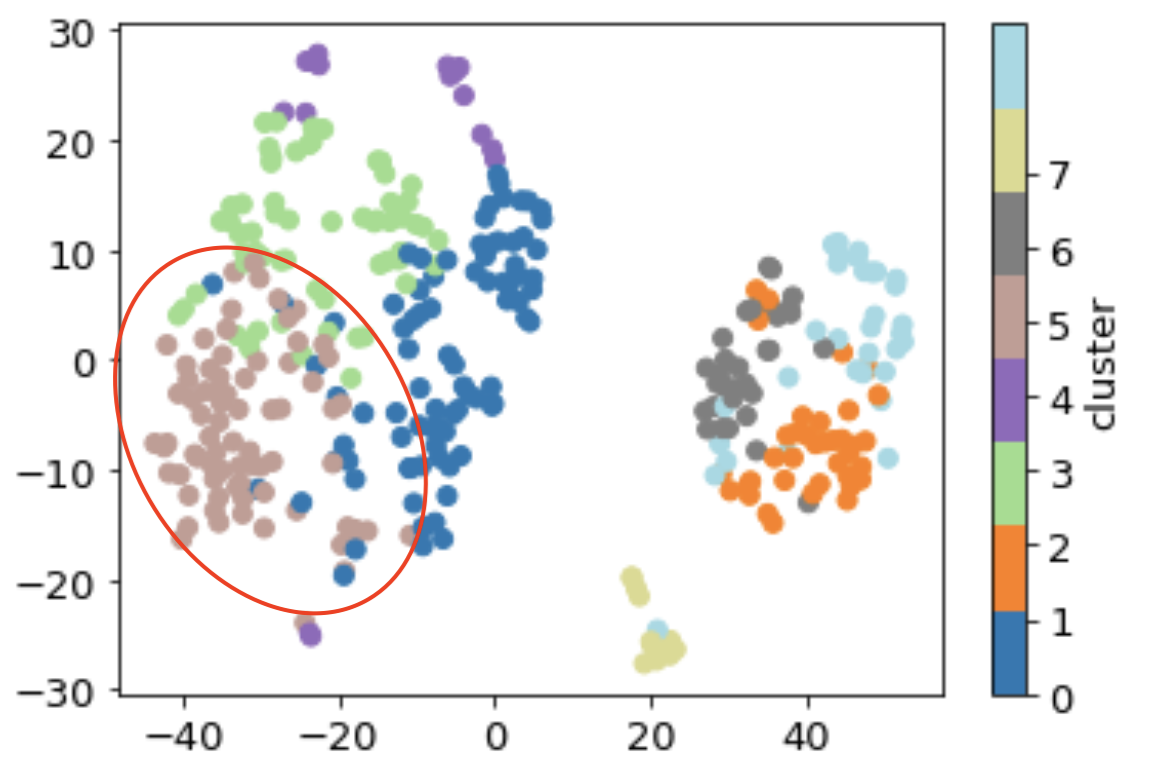}
    \caption{t-SNE results with circled most biased cluster for \textbf{HAS DIABETES} attribute}
    \label{app:fig:diabetes}
\end{figure}

\begin{table}[!ht]
\small
\centering
\begin{tabular}{l|l} 
\hline
Group                           & Percent (\%)  \\ 
\hline
Has Negative Descriptor                    & 8.86        \\
Has Diabetes                    & 35.43        \\
Has Chronic Illness                   & 88.0          \\
Medicaid Insurance              & 7.71         \\
Medicare Insurance              & 60.86        \\
Private Insurance               & 28.0            \\
Speaks English                  & 86.57        \\
Assigned Male at Birth (AMAB)   & 56.29        \\
Assigned Female at Birth (AFAB) & 43.71        \\
Self-identifies White           & 75.14        \\
Self-Identifies Black           & 13.43        \\
AFAB + Self-Identifies Black    & 8.86        \\
\hline
\end{tabular}
\caption{Percent of attribute in the MIMIC-3 data}
\label{tbl:appendix_attributes}
\end{table}

\begin{table}[t!]
\centering
\begin{adjustbox}{width=\columnwidth}
\begin{tabular}{l|l|l|l|l|l} 
\hline
\multirow{4}{*}{Has Diabetes}                    & Method  & Inertia & SCR  & SIR  & \textbar{}Bias\textbar{}  \\ 
\cline{2-6}
                                                 & K-Means & 1.00    & 0.33 & 0.27 & 0.14                      \\
                                                 & LOGAN   & 1.00    & 0.25 & 0.17 & 0.25                      \\
                                                 & SLOGAN  & 1.12    & 0.25 & 0.15 & 0.29                      \\ 
\hline\hline
\multirow{4}{*}{Has Negative}                    & Method  & Inertia & SCR  & SIR  & \textbar{}Bias\textbar{}  \\ 
\cline{2-6}
                                                 & K-Means & 1.00    & 0.00 & 0.00 & 0.00                      \\
                                                 & LOGAN   & 0.88    & 0.20 & 0.19 & 0.20                      \\
                                                 & SLOGAN  & 0.85    & 0.20 & 0.19 & 0.37                      \\ 
\hline\hline
\multirow{4}{*}{Has Chronic Illness}             & Method  & Inertia & SCR  & SIR  & \textbar{}Bias\textbar{}  \\ 
\cline{2-6}
                                                 & K-Means & 1.00    & 0.17 & 0.25 & 0.17                      \\
                                                 & LOGAN   & 1.15    & 0.40 & 0.32 & 0.40                      \\
                                                 & SLOGAN  & 0.89    & 0.50 & 0.47 & 0.23                      \\ 
\hline\hline
\multirow{4}{*}{Is Medicaid Insurance}           & Method  & Inertia & SCR  & SIR  & \textbar{}Bias\textbar{}  \\ 
\cline{2-6}
                                                 & K-Means & 1.00    & 0.40 & 0.46 & 0.23                      \\
                                                 & LOGAN   & 0.99    & 0.20 & 0.25 & 0.20                      \\
                                                 & SLOGAN  & 0.94    & 0.20 & 0.11 & 0.76                      \\ 
\hline\hline
\multirow{4}{*}{Is Medicare Insurance}           & Method  & Inertia & SCR  & SIR  & \textbar{}Bias\textbar{}  \\ 
\cline{2-6}
                                                 & K-Means & 1.00    & 0.13 & 0.13 & 0.21                      \\
                                                 & LOGAN   & 0.91    & 0.22 & 0.22 & 0.22                      \\
                                                 & SLOGAN  & 0.87    & 0.22 & 0.16 & 0.21                      \\ 
\hline\hline
\multirow{4}{*}{Is Private Insurance}            & Method  & Inertia & SCR  & SIR  & \textbar{}Bias\textbar{}  \\ 
\cline{2-6}
                                                 & K-Means & 1.00    & 0.22 & 0.20 & 0.12                      \\
                                                 & LOGAN   & 1.18    & 0.14 & 0.12 & 0.14                      \\
                                                 & SLOGAN  & 1.12    & 0.14 & 0.10 & 0.26                      \\ 
\hline\hline
\multirow{4}{*}{Is English Speaker}              & Method  & Inertia & SCR  & SIR  & \textbar{}Bias\textbar{}  \\ 
\cline{2-6}
                                                 & K-Means & 1.00    & 0.00 & 0.00 & 0.00                      \\
                                                 & LOGAN   & 1.02    & 0.17 & 0.17 & 0.17                      \\
                                                 & SLOGAN  & 0.91    & 0.43 & 0.44 & 0.31                      \\ 
\hline\hline
\multirow{4}{*}{Assigned Male at Birth (AMAB)}   & Method  & Inertia & SCR  & SIR  & \textbar{}Bias\textbar{}  \\ 
\cline{2-6}
                                                 & K-Means & 1.00    & 0.00 & 0.00 & 0.00                      \\
                                                 & LOGAN   & 1.00    & 0.11 & 0.09 & 0.11                      \\
                                                 & SLOGAN  & 1.03    & 0.25 & 0.13 & 0.41                      \\ 
\hline\hline
\multirow{4}{*}{Assigned Female at Birth (AFAB)} & Method  & Inertia & SCR  & SIR  & \textbar{}Bias\textbar{}  \\ 
\cline{2-6}
                                                 & K-Means & 1.00    & 0.00 & 0.00 & 0.00                      \\
                                                 & LOGAN   & 1.00    & 0.11 & 0.09 & 0.11                      \\
                                                 & SLOGAN  & 1.05    & 0.13 & 0.04 & 0.39                      \\ 
\hline\hline
\multirow{4}{*}{Self-identifies White}           & Method  & Inertia & SCR  & SIR  & \textbar{}Bias\textbar{}  \\ 
\cline{2-6}
                                                 & K-Means & 1.00    & 0.14 & 0.13 & 0.14                      \\
                                                 & LOGAN   & 0.86    & 0.38 & 0.37 & 0.38                      \\
                                                 & SLOGAN  & 0.98    & 0.40 & 0.41 & 0.26                      \\ 
\hline\hline
\multirow{4}{*}{Self-identifies Black}           & Method  & Inertia & SCR  & SIR  & \textbar{}Bias\textbar{}  \\ 
\cline{2-6}
                                                 & K-Means & 1.00    & 0.40 & 0.28 & 0.20                      \\
                                                 & LOGAN   & 0.91    & 0.20 & 0.10 & 0.20                      \\
                                                 & SLOGAN  & 1.02    & 0.20 & 0.10 & 0.27                      \\ 
\hline\hline
\multirow{4}{*}{Self-identifies Black + AFAB}    & Method  & Inertia & SCR  & SIR  & \textbar{}Bias\textbar{}  \\ 
\cline{2-6}
                                                 & K-Means & 1.00    & 0.20 & 0.13 & 0.28                      \\
                                                 & LOGAN   & 1.00    & 0.20 & 0.13 & 0.20                      \\
                                                 & SLOGAN  & 0.99    & 0.60 & 0.49 & 0.35                      \\
\hline
\end{tabular}
\end{adjustbox}
\caption{Comparison between K-Means, LOGAN, and SLOGAN under each attribute type. SCR and SIR are respectively the ratio of biased clusters and ratio of biased instances. \textbar{}Bias\textbar{} is the averaged absolute bias score for these biased clusters. Results not shown in \%.}
\label{tbl:appendix_comparison}
\end{table}

\begin{table}[!t]
\centering
\small
\begin{tabular}{@{}l|l} 
\hline
Group                                             & $\Delta$ (\%)  \\ 
\hline
Private Insurance                                 & -100.0                                                 \\
Medicaid Insurance                                & 11.1                                                   \\
Medicaid Insurance                                & 51.5                                                   \\
Self-Identifies White                             & 36.5                                                   \\
Self-Identifies Black                             & N/A                                                    \\
Self-Identifies Hispanic                          & N/A                                                    \\
Self-Identifies Asian                             & N/A                                                    \\
Self-Identifies Other                             & 11.1                                                   \\
English Speaker                                   & -1.6                                                   \\
Assigned Male at Birth (AMAB)                     & -38.3                                                  \\
Has Chronic Illness, Not Diabetes                 & 37.8                                                   \\
Has Chronic Illness, Not Diabetes or Hypertension & 33.3                                                   \\
Hypertensive                                      & 11.1                                                   \\
Has Acute Illness                                 & 27.8   \\
\hline
\end{tabular}     
\caption{Percentage differences ($\Delta$, in \%) in characteristics between most and least biased cluster for HAS DIABETES attribute. A positive number means the most biased cluster has more instances of this attribute versus the least biased cluster. N/A indicates division by zero.}
\label{tbl:has_diabetes_character}
\end{table}

\begin{table}[!t]
\centering
\small
\begin{tabular}{l|l|l} 
\hline
Group & $\lambda$ & $\gamma$  \\ 
\hline
Has Diabetes                                    & -30       & 50        \\
Has Negative Descriptor                         & -20       & 0         \\
Has Chronic Illness                             & -30       & 50        \\
Medicaid Insurance                              & -70       & 30        \\
Medicare Insurance                              & -50       & 0         \\
Private Insurance                               & -70       & 40        \\
Speaks English                                  & -30       & 0         \\
Assigned Male at Birth (AMAB)                   & -10       & 60        \\
Assigned Female at Birth (AFAB)                 & 0         & 70        \\
Self-Identifies White                           & -30       & 20        \\
Self-Identifies Black                           & -20       & 60        \\
AFAB + Self-Identifies Black                    & -10       & 60      \\
\hline
\end{tabular}
\caption{Hyper parameter search for $\lambda$ and  $\gamma$ after searching between combinations between -100-0 and 0-100, respectively.}
\label{tbl:appendix_params}
\end{table}

\begin{table}[!t]
\centering
\small
\begin{tabular}{@{}l|l@{}}
\hline
\multirow{5}{*}{\shortstack{Most biased  \\(38.7)}} & denies, rehab, treat, pain, well, \\
& sputum, transferred, hx, valve, \\
& sent, course, cxr, chest pain, one, \\
& episodes, mild, cough, floor, \\
& worsening, disease, tobacco            \\ 
\hline
\multirow{5}{*}{\shortstack{Least biased \\(0.67)}}   & pain, given, denies, admit, home,\\
&  time, last, well, hip,  past, started,\\
&  disease, found, noted, transferred, \\
&  liver, developed, treat, \\
& symptoms, nausea, blood  \\
\hline
\end{tabular}
\caption{Most and Least Biased LDA top 20 words for \textbf{HAS NEGATIVE DESCRIPTOR} patient descriptor. Number in parentheses is the bias score (\%) of that cluster.}
\label{tbl:lda_negative}
\end{table}

\begin{table}[t]
\centering
\small
\begin{tabular}{c|l}
\cline{1-2}
\multirow{5}{*}{\shortstack{Most biased  \\(32.7) }}  & disease, cardiac, lives, received, \\
& given, admit, denies, parent, family,  \\
& cath,  symptoms, cancer, positive,\\
&  diabetes mellitus, type,  past, time, \\
& alcohol, cad, recently, ct \\ 
\cline{1-2}
\multirow{5}{*}{\shortstack{Least biased \\ (3.4)}}  & abdominal pain, denies, pain, started, \\
& chest pain, chronic, cough, disease, \\
& transferred, past, hyperlipidemia, patient,\\
& time, given, hypert, recent, cardiac, ros, \\
& shortness breath, complaints, found          \\ 
\cline{1-2}
\end{tabular}
\caption{Top 20 topic words in the most and least biased
cluster using SLOGAN under \textbf{IS ENGLISH SPEAKER}. Number
in parentheses is the bias score (\%) of that cluster.}
\label{tbl:lda_medicaid}
\end{table}





\bibliography{aaai23}

\end{document}